\documentclass{article}

\usepackage{microtype}
\usepackage{graphicx}
\usepackage{subcaption}
\usepackage{booktabs} 

\usepackage{hyperref}

\usepackage[accepted]{icml2026}

\usepackage{amsmath}
\usepackage{amssymb}
\usepackage{mathtools}
\usepackage{amsthm}

\usepackage{url}            
\usepackage{booktabs}       
\usepackage{amsfonts}       
\usepackage{nicefrac}       
\usepackage{microtype}      
\usepackage{xcolor}         
\usepackage{graphicx}
\usepackage{multirow}
\usepackage{amsmath}
\usepackage{amssymb}
\usepackage{makecell}
\usepackage{multicol}
\usepackage{comment}
\usepackage{wrapfig}
\usepackage{xspace}
\usepackage{enumitem}

\usepackage[capitalize,noabbrev]{cleveref}

\theoremstyle{plain}

\theoremstyle{definition}

\theoremstyle{remark}

\usepackage[textsize=tiny]{todonotes}
\newcommand{\method}{{\it GeoMoLa}\xspace}
\icmltitlerunning{\method: 
Geometry-Aware Motion Latents for Learning Robust Manipulation Policies}

\begin{document}

\twocolumn[
  \icmltitle{\method: 
Geometry-Aware Motion Latents for Learning Robust Manipulation Policies}

  \icmlsetsymbol{equal}{*}

  \begin{icmlauthorlist}
    \icmlauthor{Yunchao Zhang}{yyy}
    \icmlauthor{Yijia Weng}{sch}
    \icmlauthor{Ruizhe Liu}{yyy}
    \icmlauthor{Ming Hu}{comp}
    \icmlauthor{Leonidas Guibas}{sch}
    \icmlauthor{Yanchao Yang}{yyy}
  \end{icmlauthorlist}

  \icmlaffiliation{yyy}{Department of Computer Science, The University of Hong Kong, Hong Kong, China}
  \icmlaffiliation{sch}{Department of Computer Science, Stanford University, Stanford, California, USA}
  \icmlaffiliation{comp}{Department of Data Science and AI, Monash University, Melbourne, Australia}

  \icmlcorrespondingauthor{Yanchao Yang}{ycyang@cs.hku.hk}

  \icmlkeywords{Robotics, Manipulation, Motion Planning, Deep Learning, ICML}

  \vskip 0.3in
]



\printAffiliationsAndNotice{}  

\begin{abstract}
  Learning motion latents for robotic manipulation heavily relies on extracting motion patterns from visual sequences, yet effective action abstractions require understanding three-dimensional geometric transformations. Here, we introduce \method (Geometry-Aware Motion Latents), which learns discrete motion latent codes by predicting how point clouds evolve during manipulation rather than reconstructing visual observations. This four-dimensional objective -- spatial geometry changing through time -- forces latent representations to encode actual physical motion rather than appearance patterns. \method achieves state-of-the-art performance using only single-view RGB-D input, while existing methods require multi-view reconstruction, succeeding across diverse manipulation benchmarks. Our ablations reveal that geometric prediction is the key to driving performance, quantitatively validating that manipulation depends on spatial understanding. 
Furthermore, the learned codes exhibit effective motion abstraction: applying them to novel scenes produces physically consistent transformations regardless of visual context. Our real-world experiments also confirm this robustness capability, achieving robust manipulation with minimal demonstrations in cluttered environments where geometric reasoning determines success. Thus, we demonstrate that effective motion latents for robot control can better emerge from understanding motion through its three-dimensional effects rather than pixel-level patterns.
\end{abstract}
\section{Introduction}

Robot manipulation requires learning reusable motion patterns -- 
motion latents \citep{bruce2024genie,parkerholder2024genie2,genie3} -- 
that abstract complex continuous movements into discrete, transferable skills. 
Current methods mainly learn these motion latents from sequences of two-dimensional images, missing the three-dimensional geometric structure that fundamentally determines manipulation success. 
A grasping action, for instance, depends not only on visual appearance but also on precise spatial relationships, approach angles, and the continuous evolution of three-dimensional configurations over time. 

{\it Therefore,} 
learning motion latents without access to this underlying spatiotemporal geometry may produce representations that fail to generalize across different viewpoints, object poses, or spatial arrangements.
{\it Furthermore,} this representational gap could create cascading failures in real-world deployment. 
Robots may not recognize that occluded objects maintain their geometric relationships despite visual changes, 
and small spatial errors compound across action sequences without understanding of three-dimensional workspace dynamics. 
Solving this essential representation issue in motion latent learning is necessary for robots to understand manipulation through spatial relationships and physical transformations rather than pixel patterns.

The {\it core} challenge lies in jointly modeling spatial geometry and temporal dynamics without prohibitive computational cost.
Existing methods capture either spatial structure through static three-dimensional representations \citep{Ke20243DDA,Ze20243DDP} or temporal patterns through two-dimensional video \citep{Ye2024LatentAP,Chen2024MotoLM}, 
but not both. 
Three-dimensional approaches process frozen point clouds without modeling evolution; diffusion policies generate trajectories from fixed scene features; video-based latent learning operates in image space without depth. 

\begin{figure*}[!t]
    \includegraphics[width=\textwidth]{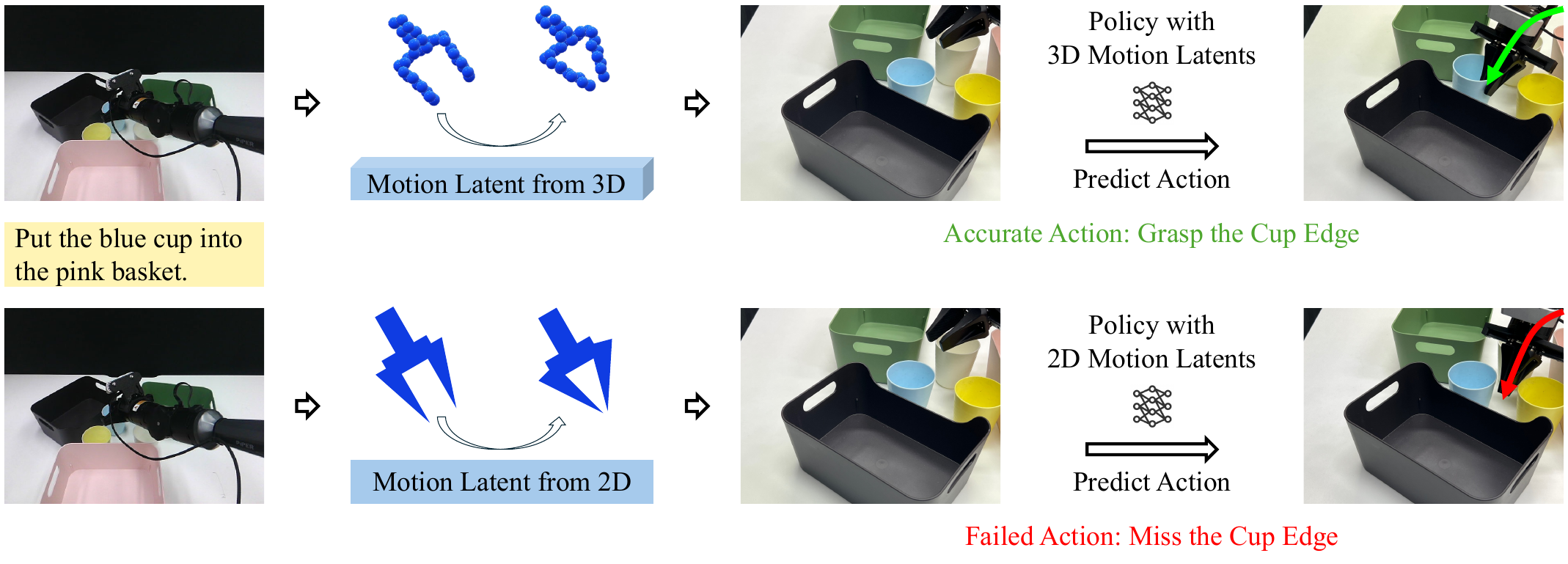}
    \vspace{-20pt}
    \caption{\textbf{
    Policies trained with 2D and 3D motion latents.} 
    When encountering novel and cluttered scenes (left) that differ from the training distribution, the policy trained with 3D motion latents (top) demonstrates superior task performance with more robust control, e.g., enhanced reaching and grasping accuracy.}
    \label{fig:teaser}
    \vspace{-10pt}
\end{figure*}

Our {\it key} insight is that effective motion latents must encode geometric transformations in three-dimensional space over time, not static scenes or visual motion. 
In this paper, we propose \method (Geometry-Aware Motion Latents), which learns {3D discrete motion latent codes (i.e., motion latent) by predicting how 3D point clouds or point maps evolve during manipulation}. 
By training latent representations to forecast future geometric states rather than reconstruct current observations, 
we ensure these codes capture the causal relationship between actions and their spatial effects, creating motion primitives grounded in physical transformation rather than visual appearance.

Our approach demonstrates that explicitly modeling four-dimensional dynamics through geometry-aware motion latents significantly improves manipulation performance. 
We achieve state-of-the-art results using only single-view RGB-D input, while our learned latent codes produce consistent geometric transformations across different scenes, validating their transferability. 
Ablation studies reveal that three-dimensional geometric prediction is critical while visual appearance contributes minimally, suggesting rethinking representation learning for manipulation. 
In real-world experiments with limited demonstrations, our method excels particularly in cluttered environments where spatial reasoning determines success. 
While focused on rigid-body manipulation, our framework establishes four-dimensional geometric prediction as an effective objective for motion latent learning for robotics.

In summary, we make the following contributions:
\begin{itemize}[left=0em, itemsep=0.5pt, parsep=0pt]
    \item We introduce the first framework that explicitly models robot manipulation as continuous four-dimensional processes while learning {geometry-aware motion latents} through self-supervised prediction of future three-dimensional point cloud evolution.
    \item We demonstrate through comprehensive ablations that geometric structure in the motion latents is significantly more important than visual appearance for manipulation success.
    \item We achieve state-of-the-art performance across multiple benchmarks while showing that learned motion latents transfer consistently across different scenes and configurations.
    \item We show superior real-world performance with limited demonstrations, particularly excelling in cluttered scenarios requiring precise spatial reasoning.
\end{itemize}

\section{Related Work}

\textbf{Motion latents.} Learning motion latent representations has proven effective across diverse applications. 
GENIE \citep{bruce2024genie,parkerholder2024genie2,genie3} maps user inputs to latent spaces for generating interactive environments, while ILPO \citep{edwards2019imitating} employs motion latents for pretraining video game policies. 
Recent work has explored deriving motion latents directly from observations: LAPA \citep{Ye2024LatentAP} and Moto \citep{Chen2024MotoLM} extract motion latents from raw inputs to leverage unlabeled data at scale. However, these observation-based approaches overlook the inherent 4D spatiotemporal structure of robotic actions. In contrast, we derive geometry-aware motion latents from 4D trajectories that explicitly capture spatial deformations and temporal dynamics, enabling more robust transfer to robotic manipulation tasks.

\textbf{Diffusion models in robotics.} Diffusion models have emerged as powerful tools for robotic manipulation, particularly for trajectory generation and action prediction. ChainedDiffuser \citep{xian2023chaineddiffuser} replaces traditional motion planners with a trajectory diffusion model that conditions on 3D scene features and predicted keyposes from Act3D \citep{Gervet2023Act3D3F} to generate linking trajectories. 
Building on this, 3D Diffuser Actor \citep{Ke20243DDA} tackles the more challenging task of jointly predicting the next keyposes and linking trajectories, while 3D Diffusion Policy \citep{Ze20243DDP} combines 3D representations with diffusion objectives. We also evaluate against recent diffusion-based methods \citep{Bu2024ClosedLoopVC,Black2023ZeroShotRM} and demonstrate superior performance. Unlike these approaches that primarily focus on static 3D representations, our method explicitly models 4D dynamics through motion latents, enabling better temporal reasoning and generalization.

\textbf{2D and 3D scene representations for robot manipulation.}  
3D scene-to-action policies address this limitation through explicit geometric reasoning: C2F-ARM \citep{James2021CoarsetoFineQE} and PerAct \citep{Shridhar2022PerceiverActorAM} voxelize workspaces but face computational scaling challenges; Act3D \citep{Gervet2023Act3D3F} avoids voxelization by sampling and featurizing 3D points through cross-attention; and RVT \citep{Goyal2023RVTRV} reprojects RGB-D inputs to multiple views before lifting predictions to 3D. End-to-end image-to-action models like RT-1 \citep{Brohan2022RT1RT}, RT-2 \citep{Brohan2023RT2VM}, GATO \citep{Reed2022AGA}, BC-Z \citep{Jang2022BCZZT}, RT-X \citep{Padalkar2023OpenXR}, Octo \citep{Team2024OctoAO}, and InstructRL \citep{Liu2022InstructionFollowingAW} directly predict 6-DoF poses from 2D images but require thousands of demonstrations to implicitly learn 3D geometry. While these methods improve upon 2D approaches through explicit 3D representations, they still treat manipulation as static spatial reasoning. Our approach advances beyond static 3D by modeling actions as continuous 4D processes, capturing how spatial configurations evolve over time.
\section{Method}
\label{sec:method}

\paragraph{Overview.}
We aim to learn robotic manipulation policies that map RGB-D observations and task instructions to executable actions. Our {\it key} contribution is learning discrete motion latents -- {\it abstract motion concepts that guide high-level policy planning} -- from 4D spatiotemporal data (3D pointmaps over time) rather than 2D video sequences.
Motion latents are essential for abstracting reusable motion patterns and enabling task generalization,
yet most existing methods learn motion latents directly from 2D observations, missing crucial 3D geometric information (depth relationships, spatial arrangements, and object poses) that fundamentally determine manipulation feasibility.

Our geometry-aware motion latents address this limitation by encoding abstract motion concepts grounded in 3D geometry.
This approach provides three advantages: better generalization across manipulation tasks through abstract motion primitives, improved interpretability by encoding 3D geometric transformations, and enhanced performance in cluttered environments requiring precise spatial reasoning. Our framework comprises two components: a self-supervised pipeline that discovers geometry-aware motion latents by predicting future 3D observations from demonstrations, and a diffusion-based model that leverages these 3D action latents to generate actions.

\subsection{Problem formulation}
We consider a dataset of robotic manipulation demonstrations
$\mathcal{D} = \{(\{\mathbf{o}_i^t, \mathbf{a}_i^t\}_{t=1}^{T_i}, l_i)\}_{i=1}^N$, where each demonstration contains a sequence of observation-action pairs with a natural language instruction $l_i$. Each observation $\mathbf{o}_i^t = (\mathbf{I}_i^t, \mathbf{D}_i^t)$ consists of an RGB image $\mathbf{I}_i^t \in \mathbb{R}^{H \times W \times 3}$ and a depth map $\mathbf{D}_i^t \in \mathbb{R}^{H \times W}$. Each action $\mathbf{a}_i^t = (\mathbf{p}_i^t, \mathbf{r}_i^t, g_i^t)$ specifies the end-effector position $\mathbf{p}_i^t \in \mathbb{R}^3$, rotation $\mathbf{r}_i^t \in \mathbb{R}^6$, and gripper state $g_i^t \in \{0,1\}$. We employ the 6D rotation representation proposed by \citet{Zhou2018OnTC} to circumvent the discontinuity issues inherent in quaternion representations.
Our objective is to learn a policy $\pi$ that maps the observation history and task instruction to an action chunk:
$\hat{\mathbf{a}}^{t:t+h-1} = \pi(\mathbf{o}^{t}, l)$,
where $h$ is the prediction horizon.
The challenge lies in capturing both geometric manipulation constraints (e.g., collision avoidance, grasp stability) and semantic language intent.
{We address this through a two-stage method:
First,
we learn a dictionary of geometry-aware motion latents (i.e., motion latents)
that capture reusable motion patterns from demonstrations
(Sec.~\ref{subsec:latent-action}).
Second,
we leverage these learned motion latents to guide a diffusion-based policy for generating executable trajectories (Sec.\ref{subsec:action-prediction}).
We now detail each component.}

\subsection{Learning {motion latents}}
\label{subsec:latent-action}

We propose a self-supervised framework that discovers motion patterns by learning a dictionary of discrete 3D latent codes, referred to as {motion latents}, encoded from the current observation and task instructions, and trained to predict future 3D observations.
We predict the future 3D geometry to encourage the motion latents to capture the underlying motions driving scene evolution. 
We use discrete rather than continuous representations because they naturally cluster similar motions into reusable primitives and provide interpretable, composable action abstractions.


\vspace{-1mm}
\subsubsection{Encoding observations into 3D motion latent}

Given a current observation 
$\mathbf{o}^t = (\mathbf{I}^t, \mathbf{D}^t)$ and task instruction $l$, 
we first extract a continuous representation using a vision-language encoder $\phi^{\text{vlm}}$ based on Mini-GPT~\citep{Zhu2023MiniGPT4EV}, which provides strong visual-linguistic grounding:
\(
\mathbf{f}^t = \phi^{\text{vlm}}(\mathbf{o}^t, l)
\).
We then discretize this representation using Vector Quantization (VQ-VAE)~\citep{Oord2017NeuralDR}, where we map $\mathbf{f}^t$ to a sequence of $n_s$ discrete codes from a learned codebook $\mathcal{C} = \{\mathbf{c}_1, \mathbf{c}_2, \ldots, \mathbf{c}_K\}$ with vocabulary size $K$:
\begin{equation}
\mathbf{z}^t = \text{VQ}(\mathbf{f}^t) = [z_1^t, z_2^t, \ldots, z_{n_s}^t],
\end{equation}
where each $z_j^t \in \{1, 2, \ldots, K\}$ is an index into the codebook. 
We employ NSVQ~\citep{vali2022nsvq} for stable training, addressing gradient collapse issues common in VQ-VAE.

\begin{figure*}[!t]
    \centering
    \includegraphics[width=\textwidth]{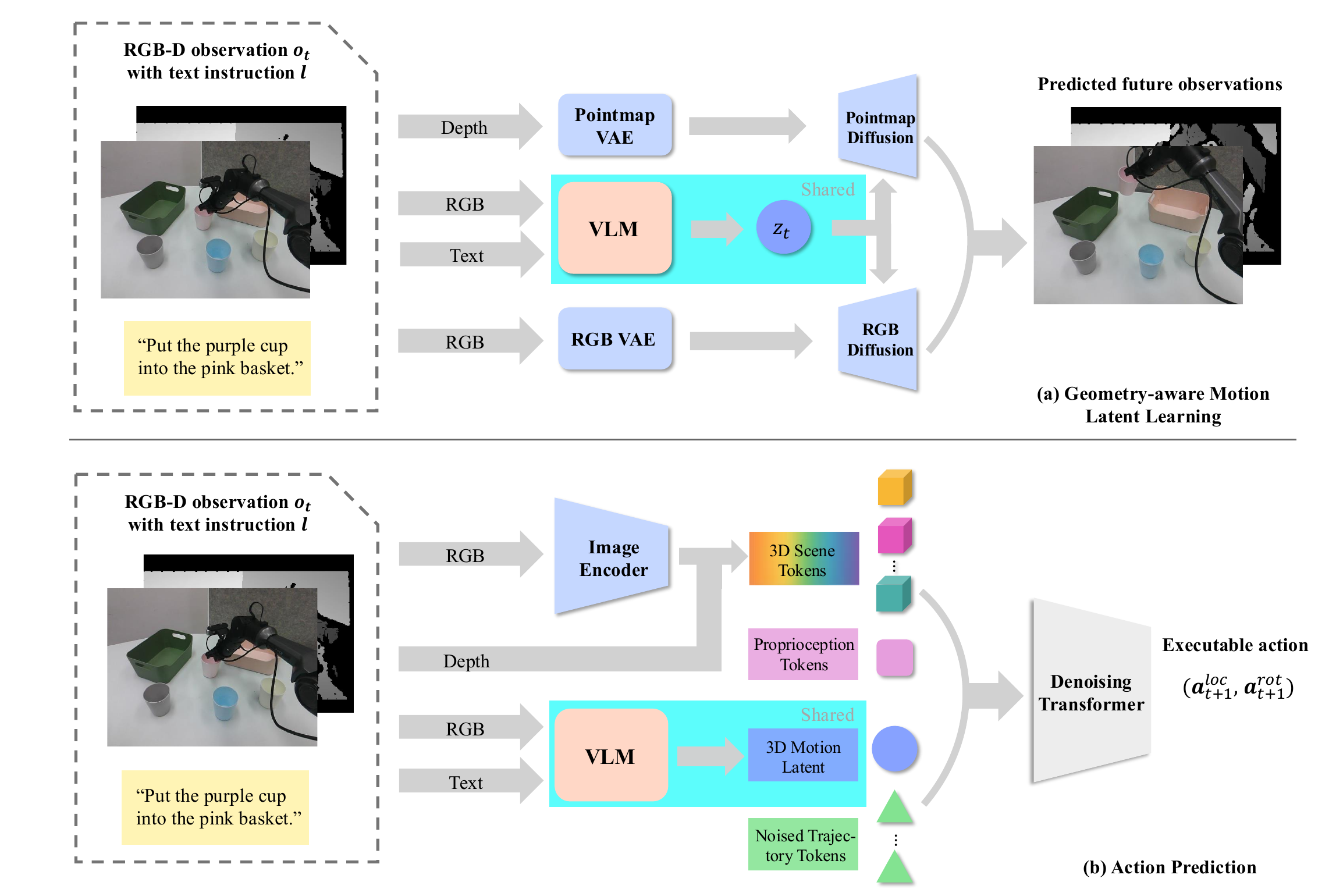}
    \caption{\textbf{\method framework.} 
    \textbf{(a) {Geometry-Aware Motion Latent Learning:}} RGB-D observations and language instructions are encoded into discrete motion latents via VQ-VAE, trained by predicting future pointmaps and RGB images. This self-supervised objective ensures latent codes capture 4D dynamics (3D geometry over time).
    \textbf{(b) Latent-Conditioned Action Prediction:} {The previously trained motion latent encoder and the codebook are applied to guide a 3D denoising transformer to generate 6-DoF trajectories through iterative refinement}, using 3D-aware attention mechanisms to accommodate geometric constraints.}
    \label{fig:method}
\end{figure*}

\subsubsection{Learning via future 3D prediction}

The core challenge in learning motion latents is ensuring they encode motion semantics rather than visual appearance. 
By requiring latent codes to predict future geometric transformations, we force them to capture the causal relationship between actions and their effects in 3D space.
Specifically, we train the latent codes $\mathbf{z}^t$ through a conditional prediction task: given the current 3D scene and a latent code, predict how the scene evolves over future timesteps. 
We use 3D pointmaps to ensure the codes learn geometric transformations rather than pixel-level appearance changes.

\textbf{Pointmap Representation.} 
We convert each RGB-D observation $\mathbf{o}^t$ to a pointmap $\mathbf{P}^t \in \mathbb{R}^{H \times W \times 3}$ by back-projecting pixels to 3D coordinates using camera intrinsics and extrinsics.

\textbf{Conditional Diffusion for Future Prediction.} 
We then employ a video latent diffusion architecture~\citep{Blattmann2023StableVD} adapted for pointmaps. 
The model consists of two components:

(1)
A 3D-aware VAE with encoder $\psi^{\text{enc}}$ and decoder $\psi^{\text{dec}}$ that translates pointmaps to and from a latent space:
\(
\mathbf{h}_{\text{pm}}^{t'} = \psi^{\text{enc}}(\mathbf{P}^{t'}), \
\hat{\mathbf{P}}^{t'} = \psi^{\text{dec}}(\hat{\mathbf{h}}_{\text{pm}}^{t'})
\),
where $\mathbf{h}_{\text{pm}}^{t'} \in \mathbb{R}^d$ denotes the latent encoding at timestep $t'$. The VAE is initialized from a pre-trained RGB VAE~\citep{Blattmann2023StableVD} and fine-tuned on pointmaps through reconstruction. See Appendix~\ref{app:imple} for details.

(2)
A diffusion model $\psi^{\text{diff}}$ that generates future latents conditioned on history and the motion latent $\mathbf{z}^t$, following DDPM~\citep{Ho2020DenoisingDP}:
\begin{equation}
\hat{\mathbf{h}}_{\text{pm}}^{t+1:t+w} = \psi^{\text{diff}}(\mathbf{h}_{\text{pm}}^{t-w+1:t}, \mathbf{z}^t; \phi),
\end{equation}
where $w$ is the observation window size. During training, we minimize the denoising objective:
\(
\mathcal{L}_{\text{diff}}^{\text{pm}} = \mathbb{E}_{k, \epsilon} \left[ \|\epsilon - \epsilon_{\phi}(\mathbf{h}_k, k, \mathbf{h}_{\text{pm}}^{t-w+1:t}, \mathbf{z}^t)\|^2 \right]
\),
where $\mathbf{h}_k = \sqrt{\alpha_k}\cdot\mathbf{h}_{\text{pm}}^{t+1:t+w} + \sqrt{1-\alpha_k}\cdot\epsilon$ is the noised latent at diffusion step $k$, $\epsilon \sim \mathcal{N}(0, I)$, and $\alpha_k$ follows the noise schedule.

We jointly train an RGB prediction branch using the same architecture and motion latent $\mathbf{z}^t$ to capture correlated appearance changes, with denoising objective $\mathcal{L}_{\text{diff}}^{\text{rgb}}$.
The combined training objective is:
\(
\mathcal{L}_{\text{total}} = \mathcal{L}_{\text{diff}}^{\text{pm}} + \mathcal{L}_{\text{diff}}^{\text{rgb}} + \mathcal{L}_{\text{vq}}
\),
where $\mathcal{L}_{\text{vq}} = \|\text{sg}[\mathbf{f}^t] - \mathbf{c}\|_2^2 + \beta\|\mathbf{f}^t - \text{sg}[\mathbf{c}]\|_2^2$ is the VQ-VAE loss with stop-gradient operator sg[·] and commitment coefficient $\beta$.
Through this objective, latent codes encoding similar motions converge to similar discrete values, creating a learned vocabulary of reusable, geometry-aware motion primitives that capture the essential dynamics of state transitions.

\subsection{Action prediction with motion latents}
\label{subsec:action-prediction}

{Having learned motion latents through self-supervised future prediction},
We now describe how to use learned geometry-aware motion latents to generate executable robot trajectories. 
The motion latent $\mathbf{z}^t$ serves as a bridge between high-level task understanding and low-level control, abstracting the essential motion while filtering out irrelevant visual details.

Our proposed action prediction model is a conditional diffusion policy that generates 6-DoF end-effector trajectories given: 
(1) the current scene observation, (2) the learned motion latent code, and (3) the robot's proprioceptive state. 
By conditioning on geometry-aware motion latents rather than raw language instructions or pixels,
the policy benefits from motion priors learned across the entire dataset, 
improving both sample efficiency and generalization.

\subsubsection{Overall Pipeline}
Specifically, we employ a 3D denoising transformer $\epsilon_\theta$ based on~\citet{Ke20243DDA} to generate executable trajectories through conditional diffusion.
The model iteratively refines noisy action sequences into precise robot motions.
Given the current observation $\mathbf{o}^t$, motion latent $\mathbf{z}^t$, and proprioceptive state $\mathbf{c}^t$,
the model generates an action chunk $\mathbf{a}^{t:t+h-1}$ over a horizon $h$ via DDPM,
where each action $\mathbf{a}^{t+k} = (\mathbf{p}^{t+k}, \mathbf{r}^{t+k}, g^{t+k})$ specifies the end-effector position, rotation, and gripper state.

\textbf{Input Tokenization.} The transformer processes four types of tokens (illustrated in Fig.~\ref{fig:method}):
\textit{\textbf{(1) Motion Latent Tokens}}: The discrete motion latent $\mathbf{z}^t = [z_1^t, \ldots, z_{n_s}^t]$ from Sec.~\ref{subsec:latent-action} is embedded via learned embeddings $\mathbf{E}_z \in \mathbb{R}^{n_s \times d_{\text{code}}}$, providing high-level motion guidance.
\textit{\textbf{(2) Trajectory Tokens}}: Each noisy action is encoded through an MLP to produce token $\mathbf{t}_{\text{traj}}^k \in \mathbb{R}^{d_{\text{model}}}$. We further concatenate the 3D position $\mathbf{p}^{t+k}$ as positional information to maintain spatial grounding.
\textit{\textbf{(3) Scene Tokens}}: We extract visual features $\mathbf{F} \in \mathbb{R}^{H \times W \times 3}$ using a frozen CLIP-ResNet50~\citep{Radford2021LearningTV} encoder. Each feature $\mathbf{F}_{ij}$ at spatial location $(i,j)$ is lifted to 3D position $\mathbf{q}_{ij}$ using the depth map and camera intrinsic matrix. This produces $H \times W$ scene tokens $\{(\mathbf{F}_{ij}, \mathbf{q}_{ij})\}$ combining appearance and 3D position.
\textit{\textbf{(4) Proprioception Token}}: The robot state $\mathbf{c}^t$ (joint angles, end-effector pose) is encoded as $\mathbf{t}_{\text{prop}} = \text{MLP}(\mathbf{c}^t) + \text{PosEmbed}(\mathbf{p}_{\text{ee}}^t)$, where $\mathbf{p}_{\text{ee}}^t$ is the current end-effector position and $\text{PosEmbed}$ denotes the positional embedding.

\textbf{Attention Mechanisms.} Then, the transformer employs a two-stage attention strategy to integrate spatial, temporal, and task information:
\textit{\textbf{(1) Self-Attention with 3D Positional Encoding}}: We first apply self-attention across all trajectory, scene, and proprioception tokens. To encode spatial relationships, we use the rotary positional embeddings~\citep{Su2021RoFormerET}.
\textit{\textbf{(2) Cross-Attention to Motion Latents}}: After self-attention, we apply cross-attention from trajectory, scene and proprioception tokens to the motion latent embeddings.
This mechanism allows the geometry-aware motion patterns encoded in $\mathbf{z}^t$ to guide trajectory generation. 
Unlike conditioning on raw language, the discrete motion latents provide structured, geometrically consistent motion priors based on real-time observation.

Finally, we apply MLPs to predict the noise added to the sequence of 3D translations $\epsilon_{\theta,p}$ and rotations $\epsilon_{\theta,r}$, as well as the gripper state $\hat{g}$, with the final trajectory tokens from the transformer outputs. This progressively refines the action estimate based on scene geometry and learned motion patterns.

\subsubsection{Training objective and inference process for action prediction}

\textbf{To learn,} 
we train the denoising transformer $\epsilon_\theta$ to predict the noise added to the ground-truth action chunk. 
Given a clean action chunk $\mathbf{a}^{t:t+h-1}$
from demonstrations, we sample noise $\boldsymbol{\epsilon} \sim \mathcal{N}(0, I)$ at diffusion step $i$, then create the noisy action chunk:
\(
\mathbf{a}_i = \sqrt{\bar{\alpha}_i}\cdot\mathbf{a}^{t:t+h-1} + \sqrt{1-\bar{\alpha}_i}\cdot\boldsymbol{\epsilon}
\).
The model learns to predict $\boldsymbol{\epsilon}$ given the noisy action chunk and conditioning:
\(
\mathcal{L}_\theta = \mathbb{E}_{i, \boldsymbol{\epsilon}} \left[\|\boldsymbol{\epsilon} - \epsilon_\theta(\mathbf{a}_i, i, \mathbf{o}^t, \mathbf{z}^t, \mathbf{c}^t)\| \right]
\).
Since actions have components with different scales and properties, we use component-specific losses:
\begin{align}
\|\boldsymbol{\epsilon} - \epsilon_\theta(\cdot)\| = &\lambda_p \|\boldsymbol{\epsilon}_p - \epsilon_{\theta,p}(\cdot)\|_1 +  \lambda_r \|\boldsymbol{\epsilon}_r - \epsilon_{\theta,r}(\cdot)\|_1 \nonumber \\ 
 &+ \lambda_g \cdot \text{BCE}(g, \hat{g})
\end{align}
where subscripts $p, r, g$ denote position, rotation, and gripper components, respectively. 
We use L1 loss for continuous values (which is more robust to outliers in trajectory data) and binary cross-entropy (BCE) for the discrete gripper state. The weights $\{\lambda_p, \lambda_r, \lambda_g\}$ are determined with tuning.

\textbf{For inference,} we sample a noisy action chunk $\mathbf{a}_N \sim \mathcal{N}(0, I)$ and iteratively denoise it following DDPM \citep{Ho2020DenoisingDP} using the learned model and the current observation's motion latent.
The final denoised action chunk contains executable 6-DoF poses and gripper commands that can be directly sent to the robot controller. 
The geometry-aware motion latent $\mathbf{z}^t$ ensures that the generated trajectory respects both the task semantics and 3D spatial constraints learned from demonstrations.

\begin{table*}[h]
\caption{\textbf{Task success rates on RLBench (single view).} \method\ achieves the highest average performance (84.7\%) and ranks first on 8 out of 10 tasks. Results averaged over $5$ seeds.}
\centering
\resizebox{\textwidth}{!}{
\begin{tabular}{lccccccccccc}
\toprule
Method & {\makecell{Close\\jar}} & {\makecell{Open\\drawer}} & {\makecell{Sweep to\\dustpan}} & {\makecell{Turn\\tap}} & {\makecell{Meat off\\grill}} & {\makecell{Stack\\blocks}} & {\makecell{Slide\\block}} & {\makecell{Put in\\drawer}} & {\makecell{Drag\\stick}} & {\makecell{Push\\buttons}} & Avg. \\
\midrule
GNFactor~\citep{Ze2023GNFactorMR} & 25.3 & 76.0 & 28.0 & 50.7 & 57.3 & 4.0 & 20.0 & 0.0 & 37.3 & 18.7 & 31.7 \\
ManiGaussian~\citep{Lu2024ManiGaussianDG} & 28.0 & 76.0 & 64.0 & 56.0 & 60.0 & 12.0 & 24.0 & 16.0 & 92.0 & 20.0 & 44.8 \\
Act3D~\citep{Gervet2023Act3D3F} & 52.0 & 84.0 & 80.0 & 64.0 & 66.7 & 0.0 & \textbf{100.0} & 54.7 & 86.7 & 64.0 & 65.3\\
SkillDiffuser~\citep{Liang2023SkillDiffuserIH} & 64.2 & 81.0 & 96.6 & 70.6 & 72.1 & 4.0 & 87.0 & 89.2 & 95.6 & 83.8 & 74.4 \\
3D Diffuser Actor~\citep{Ke20243DDA} & 66.4 & 85.6 & \textbf{98.4} & 75.2 & 76.0 & 4.0 & 87.2 & \underline{94.4} & \textbf{98.4} & 84.0 & 77.0 \\
RVT2~\citep{goyal2024rvt2} & {\underline{67.5}} & \underline{{87.2}} & {96.4} & {\underline{76.6}} & {\underline{79.6}} & {\underline{34.8}} & {86.0} & {93.2} & {97.4}  & {\underline{85.4}} & {\underline{80.4}} \\
\method\ (Ours) & \textbf{69.0} & \textbf{86.6} & \textbf{98.0} & \textbf{81.4} & \textbf{78.8} & \textbf{54.2} & \underline{92.8} & \textbf{95.2} & \underline{98.2} & \textbf{93.0} & \textbf{84.7} \\
\bottomrule
\end{tabular}
}
\label{tab:rlbench}
\vspace{-2ex}
\end{table*}

\section{Experiments}
\label{sec:exp}

We evaluate \method\ through comprehensive experiments designed to answer three key questions: 
(1) Does 4D spatiotemporal modeling improve manipulation performance compared to 2D/3D baselines? 
(2) Are the geometry-aware motion latents interpretable and transferable across tasks? 
(3) Does our approach generalize effectively to real-world scenarios with occlusion and clutter?
We conduct experiments on two simulation benchmarks (RLBench~\citep{James2019RLBenchTR} and CALVIN~\citep{Mees2021CALVINAB}) and real-world manipulation tasks using the ALOHA robot~\citep{Fu2024MobileAL}, with extensive ablations to validate our design choices.

\subsection{RLBench Evaluation}

\textbf{Benchmark description.}
RLBench is built on top of the CoppeliaSim simulator~\citep{Rohmer2013VREPAV}, using a Franka Panda robot to interact with the environment. Our model and all baselines are trained to predict the next end-effector keypose rather than the entire trajectory. To execute the predicted keypose, we use RLBench’s built-in BiRRT motion planner to generate a feasible trajectory.
For evaluation, we select a suite of 10 challenging language-conditioned manipulation tasks, including 166 variations. These variations vary in several types, like position, shape, and color. We use the front-view RGB-D camera as input to comply with practical deployment conditions. Performance is measured by task completion success rate, defined as the proportion of execution trajectories that satisfy the language-specified goal conditions.
GNFactor~\citep{Ze2023GNFactorMR} and ManiGaussian~\citep{Lu2024ManiGaussianDG} require an additional 19 views for 3D reconstruction during training. All other models receive only the front-view observation as input during inference.

\textbf{Baselines.} 
We compare against three types of methods:
\textit{(i) 3D representation-based policies:} Act3D~\citep{Gervet2023Act3D3F} voxelizes the workspace and predicts 3D action maps; 3D Diffuser Actor~\citep{Ke20243DDA} uses 3D feature fields with diffusion-based trajectory generation; GNFactor~\citep{Ze2023GNFactorMR} leverages neural radiance fields~\citep{Mildenhall2020NeRF} for scene understanding. {RVT2~\citep{goyal2024rvt2} reconstructs scene point cloud for better pose estimation.} \textit{(ii) 4D dynamic framework:} ManiGaussian~\citep{Lu2024ManiGaussianDG} employs dynamic 3D Gaussian Splatting~\citep{Kerbl20233DGS} for scene-level spatiotemporal dynamics.
\textit{(iii) motion latent methods:} SkillDiffuser~\citep{Liang2023SkillDiffuserIH} learns discrete skills from 2D observations without explicit 3D geometric grounding.

\textbf{Quantitative Results.} 
As shown in Tab.~\ref{tab:rlbench}, our method achieves the highest overall performance (84.7\%), ranking first on 8 out of 10 tasks. Unlike prior works that condition their action prediction models solely on static 3D feature fields, our approach explicitly models dynamics by learning motion latents from state transitions. SkillDiffuser also employs latent skill learning, our method further leverages 3D geometric changes along trajectories, enabling more faithful representation of scene structure and, consequently, more precise and reliable action generation. {While RVT2 utilizes RGB-D input, it obtains supervision only from ground-truth poses. Our results demonstrate that learning motion patterns enables more accurate action prediction without requiring additional views.}

\textbf{Interpretability of Geometry-Aware Motion Latents.}
To validate that our learned motion latents encode semantically meaningful motion primitives rather than task-specific behaviors, we conduct cross-scenario generalization experiments. We extract latent codes from successful trajectories and apply them to different scenes, measuring the consistency of the resulting motion patterns.

\begin{figure*}[!t]
    \includegraphics[width=\textwidth]{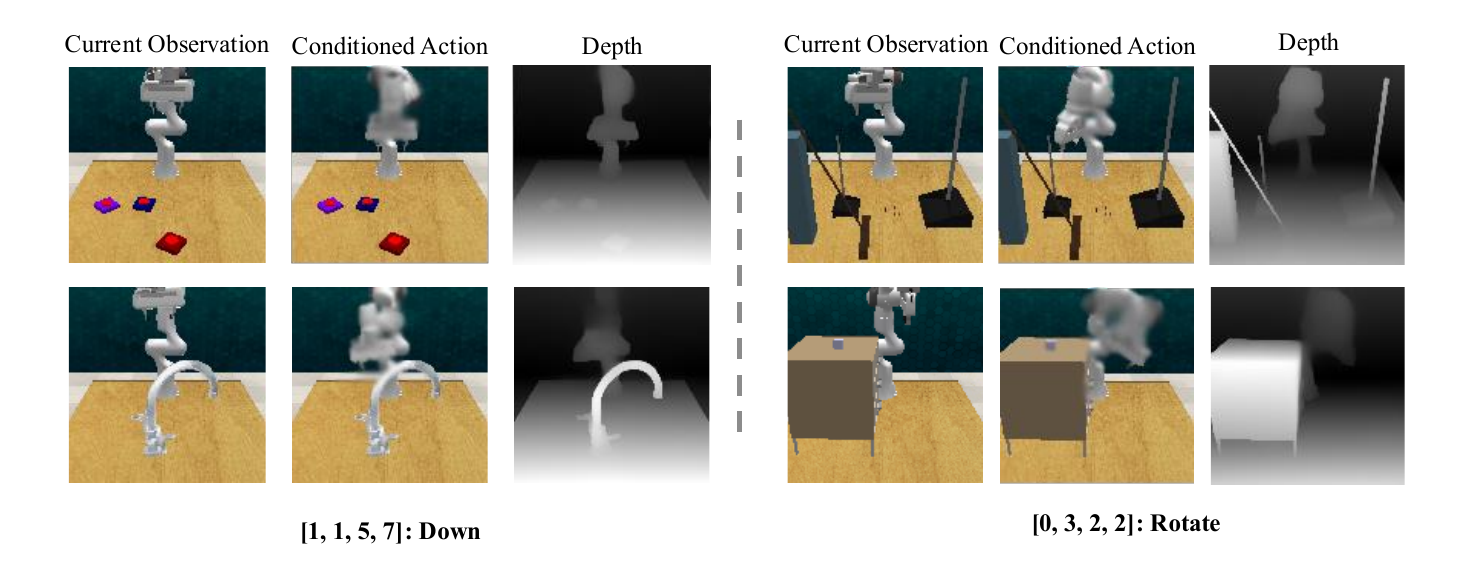}
    \vspace{-20pt}
    \caption{\textbf{Cross-scenario motion latent consistency.}
    Visualization of predicted future observations when conditioning the same latent code on different initial scenes.
    Left: Latent code $[1, 1, 5, 7]$ consistently produces downward motion.
    Right: Latent code $[0, 3, 2, 2]$ consistently generates rotational motion.}
    \label{fig:vis}
    \vspace{-1pt}
\end{figure*}

\begin{table*}[h]
\caption{\textbf{Long-horizon task chaining on CALVIN.} Methods are evaluated on completing sequences of 1-5 tasks in the unseen environment D. \method\ achieves the longest average task chain (3.60).}
\centering
\resizebox{\textwidth}{!}{
\begin{tabular}{llcccccc}
\toprule
\multirow{2}{*}{Method} & \multirow{2}{*}{Training Data}  & \multicolumn{5}{c}{Task Completion Rate (\%)} & \multirow{2}{*}{Avg. Length} \\
 \cmidrule(lr){3-7} & 
 & 1 Task & 2 Tasks & 3 Tasks & 4 Tasks & 5 Tasks & \\
\midrule
DP3~\citep{Ze20243DDP} & Language-annotated & 28.7 & 2.7 & 0.0 & 0.0 & 0.0 & 0.31 \\
GR-1~\citep{Wu2023UnleashingLV} & Language-annotated & 85.4 & 71.2 & 59.6 & 49.7 & 40.1 & 3.06 \\
SuSIE~\citep{Black2023ZeroShotRM} & All play data & 87.0 & 69.0 & 49.0 & 38.0 & 26.0 & 2.69 \\

{RVT2}~\citep{goyal2024rvt2} & {Language-annotated} &  {90.4} & {76.8} &  {61.6}	& {50.2} & {39.8} &  {3.23}\\
3D Diffuser Actor~\citep{Ke20243DDA} & Language-annotated & 93.8 & 80.3 & 66.2 & 53.3 & 41.2 & 3.35 \\
Clover~\citep{Bu2024ClosedLoopVC} & Language-annotated & \underline{96.0} & \underline{83.5} & \underline{70.8} & \underline{57.5} & \underline{45.4} & \underline{3.53}  \\
\method\ (Ours) & Language-annotated & \textbf{96.4} & \textbf{85.2} & \textbf{73.0} & \textbf{59.9} & \textbf{47.1} & \textbf{3.60} \\
\bottomrule
\end{tabular}
}
\label{tab:calvin}
\vspace{-2ex}
\end{table*}


Fig.~\ref{fig:vis} demonstrates this cross-scenario transfer capability. When latent code $[1, 1, 5, 7]$ is applied to diverse initial configurations, it consistently generates downward motion in the predicted future observations. Similarly, code $[0, 3, 2, 2]$ reliably produces rotational motion regardless of scene content. This semantic consistency validates that our 4D dynamics pretraining successfully distills reusable motion primitives from the continuous space of robot actions.

\subsection{CALVIN Evaluation}

\textbf{Benchmark Description.}
CALVIN~\citep{Mees2021CALVINAB} evaluates long-horizon task execution in PyBullet, requiring sequential completion of language-specified sub-tasks. It includes 34 distinct tasks across 4 environments (A, B, C, D) with varying textures and object positions. 
Each environment contains a Franka Emika Panda robot, desk, sliding door, drawer, LED button, light switch, and colored blocks. 
CALVIN provides 24 hours of teleoperated unstructured play data, 35\% of which are annotated with language descriptions (18k trajectory videos). Each instruction chain includes five language instructions that need to be executed sequentially. We evaluate on the challenging zero-shot generalization setting: training on environments A, B, C and testing on the unseen environment D.

\textbf{Baselines.}
We compare against: (i) \textit{3D-based methods}: DP3~\citep{Ze20243DDP} encodes RGB-D into 3D features for diffusion trajectory prediction (we add language conditioning following~\citep{Ke20243DDA}); {3D Diffuser Actor and RVT2 that previously mentioned}. (ii) \textit{Video-pretrained models}: GR-1~\citep{Wu2023UnleashingLV} leverages large-scale internet video pretraining; SuSIE~\citep{Black2023ZeroShotRM} uses all available play data including unannotated sequences. (iii) \textit{Closed-loop methods}:  Clover~\citep{Bu2024ClosedLoopVC} encodes RGB-D inputs, introducing a closed-loop error correction mechanism to achieve better video diffusion policy. 

\textbf{Results Analysis.}
Tab.~\ref{tab:calvin} shows \method\ achieves the highest average task sequence length of 3.60, indicating a higher ability when completing the entire long-horizon subtask sequences. This trend validates our hypothesis: explicit 4D dynamics modeling reduces compounding errors that accumulate over long horizons.  GR-1's internet-scale pretraining provides strong single-task performance but degrades rapidly in multi-task scenarios, suggesting that generic video understanding does not transfer directly to precise robotic control.

\begin{table*}[h]
\centering
\caption{\textbf{Real-world manipulation performance.} Success rates over 10 trials per task. \method\ shows consistent improvements, particularly in cluttered scenarios (Clean cup). † indicates tasks with significant occlusion during execution.}
\label{tab:real}
\vspace{1ex}
\resizebox{\textwidth}{!}{
\begin{tabular}{lccccccc}
\toprule
Method & Clean cup† & Stack cups & Put cups on shelf & Stack cubes & Place dish & Place cube & Average \\
\midrule
SkillDiffuser~\citep{Liang2023SkillDiffuserIH} & 30.0\% & 0.0\% & 20.0\% & 20.0\% & 20.0\% & 40.0\% & 21.7\%\\
3D Diffuser Actor~\citep{Ke20243DDA} & \underline{50.0\%} & \underline{10.0\%} & \textbf{40.0\%} & 20.0\% & 40.0\% & 80.0\% & \underline{40.0\%} \\
{GeoMoLa w/o Pointmap} & {20.0\%} & {\underline{10.0\%}} & {30.0\%} & {\underline{20.0\%}} & {\textbf{50.0\%}} & {\underline{80.0\%}} & {35.0\%} \\
\method\ (Ours) & \textbf{60.0\%} & \textbf{30.0\%} & \textbf{40.0\%} & \textbf{50.0\%} & \textbf{50.0\%} & \textbf{90.0\%} & \textbf{53.3\%} \\
\bottomrule
\end{tabular}
}
\vspace{-1ex}
\end{table*}

\subsection{Ablation Studies}

\begin{table}[ht]
\centering
\caption{\textbf{Impact of 4D dynamics learning.}}
\vspace{1ex}
\label{tab:ablation}
\resizebox{0.5\textwidth}{!}{
\begin{tabular}{lcccccc|c}
\toprule
\multirow{2}{*}{Method Variant} & \multicolumn{6}{c|}{CALVIN (Task Completion \%)} & \multicolumn{1}{c}{RLBench} \\
\cmidrule(lr){2-7} \cmidrule(lr){8-8}
 & 1 Task & 2 Tasks & 3 Tasks & 4 Tasks & 5 Tasks & Avg. Len & Success Rate \\
\midrule
 w/o Pointmap & 94.4 & 81.0 & 67.6 & 54.9 & 43.2 & 3.38 & 79.6\% \\
 w/o RGB & \underline{95.0} & \underline{82.3} & \underline{69.6} & \underline{55.2} & \underline{45.0} & \underline{3.50} & \underline{82.0\%} \\
Full & \textbf{96.4} & \textbf{85.2} & \textbf{73.0} & \textbf{59.9} & \textbf{47.1} & \textbf{3.60} & \textbf{84.7\%} \\
\bottomrule
\end{tabular}
}
\end{table}

\vspace{-1ex}
To validate our architectural choices, we conduct ablation studies examining the contribution of each modality in the 4D dynamics learning phase. We design two baselines that pretrain the motion latent space using only a single modality -- either 2D image prediction or pointmap prediction -- by removing the pointmap or RGB prediction branch from our framework, respectively.

\textbf{Importance of Geometric vs. Appearance Modeling.}
Tab.~\ref{tab:ablation} reveals a critical insight: removing pointmap prediction causes a substantial performance drop (CALVIN: -0.22 avg. length; RLBench: -5.1\%), while removing RGB prediction has minimal impact (-0.10 avg. length; -2.7\%). This asymmetry demonstrates that explicit 3D geometry is fundamental for learning transferable manipulation primitives, while appearance primarily provides auxiliary context.

\textbf{The role of geometry in different motion types.} By analysing the task-specific results on RLBench, we find that in rotation-heavy tasks (e.g., ``sweep to dustpan'', ``open drawer''), removing point map prediction reduced success rates by nearly 9\% on average, while in translation-heavy tasks (e.g., ``close jar'', ``push buttons''), the drop was less than 1\%. This suggests that 3D geometry is especially important for tasks involving fine rotational control and complex object interactions, aligning with our motivation that geometric awareness supports more precise gripper state estimation and interaction reasoning.  Thus, the key value of our approach lies in providing consistent robustness in geometrically complex settings, rather than a uniform boost across all tasks.


\subsection{Real-World Validation}

\textbf{Experimental Setup.}
We collected real-world demonstration data using the ALOHA robot platform.  RGB-D observations are captured via an Intel RealSense camera at 640×480 resolution from a front view and subsequently downsampled to 256×256 for processing. During inference, target gripper poses are executed using the MoveIt package in ROS~\citep{Coleman2014ReducingTB}.

\textbf{Data Collection and Training.}
We consider six distinct tasks, with $20$ demonstration trajectories recorded for each task. To ensure diversity within each task, variations in object quantities, positions, colors, and other attributes were intentionally introduced across different trajectories. The collected demonstration data were subsequently used to train our model. During the testing phase, the model was evaluated on the same set of tasks under zero-shot transfer conditions, where object configurations, spatial layouts, and visual properties such as color were systematically altered to assess generalization beyond the training demonstrations.  We evaluated 10 episodes for each task and reported the success rate. Models are trained from scratch on this limited data to evaluate sample efficiency in real-world settings. Detailed task information and visualization of our settings could be found in Appendix.~\ref{app:real}.

\textbf{Results and Analysis.}
Tab.~\ref{tab:real} shows \method\ achieving a 53.3\% average success rate, a 13.3\% improvement over 3D Diffuser Actor. In particular, we observe consistent gains in tasks such as ``Clean cup'', which involves significant occlusion, as well as ``Stack cups'' and ``Stack cubes'', which require precise manipulation. These improvements highlight the benefits of our approach in modeling dynamics and scene geometry, enabling more accurate action prediction.

\section{Conclusion}

This work demonstrates that motion latent learning for robotic manipulation benefits significantly from grounding in four-dimensional geometric transformations rather than visual sequences. 
Our ablation studies provide quantitative evidence: removing geometric prediction degrades motion latent quality substantially while visual prediction contributes minimally to performance. 
The success with limited real-world demonstrations indicates that geometry-aware latent representations naturally capture motion primitives without requiring extensive datasets. 
Future work could extend this motion latent framework to deformable objects and investigate hierarchical planning where high-level policies compose learned geometric primitives. 
Our results suggest that effective motion latent representations for robotics should encode how objects move through three-dimensional space over time rather than how they appear visually, providing another perspective on learning reusable manipulation skills from unlabeled demonstrations.

\section*{Acknowledgement}
This work is supported by the Early Career Scheme of the
Research Grants Council (RGC) grant \# 27207224, the
HKU-100 Award, and the HKU Shanghai Intelligent Com-
puting Research Center (ICRC).

\section*{Impact Statement}
This research advances the field of robot learning by introducing a geometry-aware framework for learning motion representations from 3D point cloud sequences. The proposed approach enables more robust and interpretable manipulation policies, particularly in cluttered and occluded environments, with potential applications in assistive robotics, healthcare, logistics, and domestic automation. By reducing reliance on large-scale labeled demonstrations, our method may lower barriers to deploying robots in diverse real-world settings. We acknowledge the importance of safety, transparency, and ethical considerations in the development and deployment of such systems, and encourage further research into human-robot collaboration, fairness, and responsible innovation in autonomous robotics.

\bibliography{example_paper}
\bibliographystyle{icml2026}

\newpage
\appendix
\onecolumn
\section{The Use of Large Language Models}
This work utilized large language models as supplementary tools to enhance writing quality, including improving clarity, maintaining consistency between sections, and refining adherence to academic writing conventions.
The core research concepts, methodological approaches, and findings represent original contributions by the authors.

\section{Simulation experiments}
\label{app:sim_exp}
\subsection{Dataset composition}

\paragraph{RLBench.} RLBench is a comprehensive and large-scale benchmark and learning environment designed to advance research in vision-guided robotic manipulation. The platform is tailored to support a variety of research areas, including reinforcement learning, imitation learning, multi-task learning, and few-shot learning.

In our specific experimental setup, we utilize a subset of 10 manipulation tasks from the RLBench environment to evaluate the multi-task capabilities of our agents. These tasks are selected to cover a diverse range of challenges, involving different objects, objectives, and required skills. The variations within each task are designed to test an agent's ability to understand and adapt to changes in color, placement, size, and object category.

Tab.~\ref{tab:rlbench_tasks} shows the details of  the composition of the  RLBench dataset.

\begin{table}[htbp]
\centering
\caption{Dataset composition of 10 manipulation tasks in RLBench \cite{James2019RLBenchTR}.}
\resizebox{\textwidth}{!}{
\begin{tabular}{llcccc}
\toprule
\textbf{Task} & \textbf{Variation Type} & \textbf{\# of Variations} & \textbf{Avg. Keyframes} & \textbf{Language Description Example} \\
\midrule
close jar & color & 20 & 6.0 & ``close the --- jar'' \\
meat off grill & category & 2 & 5.0 & ``take the --- off the grill'' \\
open drawer & placement & 3 & 3.0 & ``open the --- drawer'' \\
sweep to dustpan & size & 2 & 4.6 & ``sweep dirt to the --- dustpan'' \\
turn tap & placement & 2 & 2.0 & ``turn --- tap'' \\
slide block & color & 4 & 4.7 & ``slide the block to --- target'' \\
put in drawer & placement & 3 & 12.0 & ``put the item in the --- drawer'' \\
drag stick & color & 20 & 6.0 & ``use the stick to drag the cube onto the --- --- target'' \\
push buttons & color & 50 & 3.8 & ``push the --- button, [then the --- button]'' \\
stack blocks & color, count & 60 & 14.6 & ``stack --- --- blocks'' \\
\bottomrule
\end{tabular}
}
\label{tab:rlbench_tasks}
\end{table}

\paragraph{CALVIN.} A key evaluation protocol within CALVIN is the ``ABC$\rightarrow$D'' setup, which is specifically designed to test an agent's ability to generalize to a new, unseen environment. This setup is considered one of the most challenging evaluations in the benchmark.

\begin{figure}[!t]
    \includegraphics[width=\textwidth]{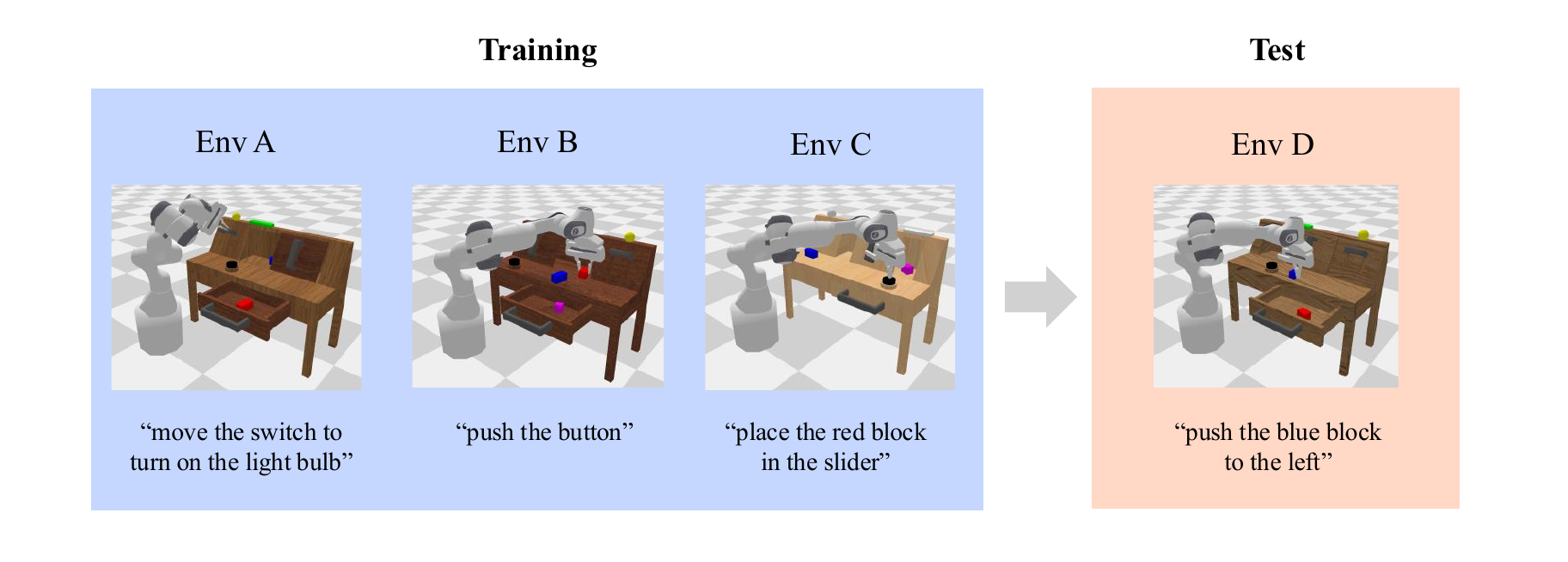}
    \vspace{-20pt}
    \caption{\textbf{llustration of the four different environments in
    CALVIN~\citep{Mees2021CALVINAB}.}}
    \label{fig:calvin_data}
    \vspace{-10pt}
\end{figure}

The visualization of the Calvin setting is demonstrated in Fig.~\ref{fig:calvin_data}. This zero-shot generalization task, where the agent must apply learned skills to a completely new setting, is crucial for assessing the robustness and adaptability of the control policy. The ABC$\rightarrow$D setup measures how well a policy can transfer its understanding of language and manipulation to an unfamiliar setting.

\subsection{Additional results}
\begin{figure*}[!t]
    \includegraphics[width=\textwidth]{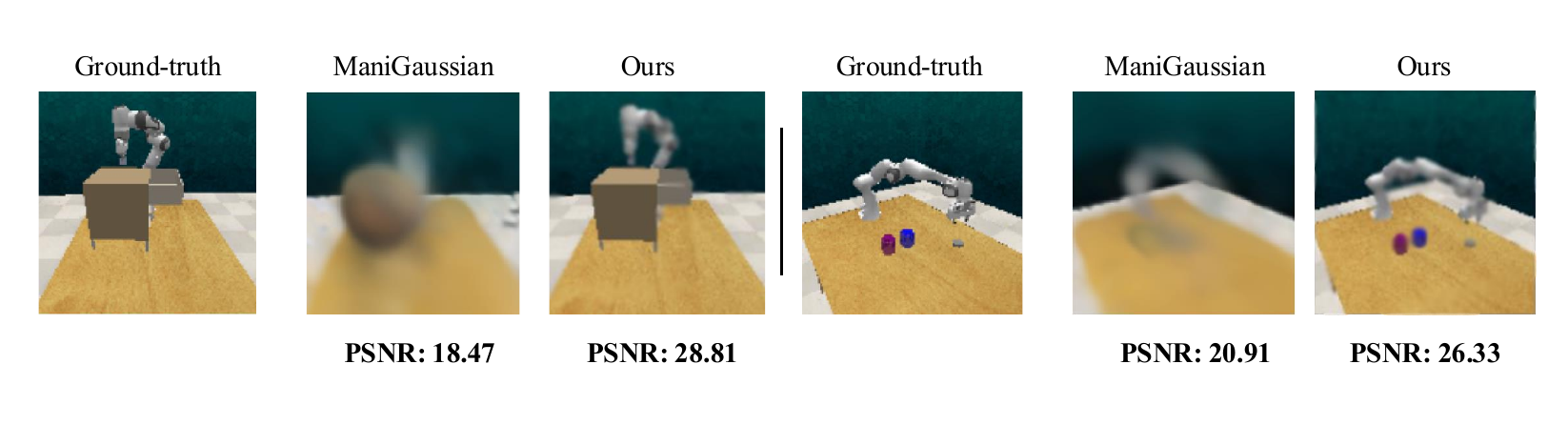}
    \vspace{-25pt}
    \caption{\textbf{Qualitative comparison of future observation prediction.}
    ManiGaussian requires 19 additional training views and produces blurred predictions with geometric inconsistencies (see distorted object boundaries).}
    \label{fig:rec}
    \vspace{-10pt}
\end{figure*}

\textbf{Future Observation Prediction Quality.}
Beyond semantic consistency, accurate future prediction is crucial for action planning. We evaluate the quality of predicted observations using both perceptual and geometric metrics.
Fig.~\ref{fig:rec} demonstrates our superior performance compared to another reconstruction-based 4D dynamic method. Despite ManiGaussian's access to 19 additional camera views for the training of dynamic Gaussian Splatting, our pointmap approach produces more accurate predictions with sharper object boundaries. The pointmap representation naturally preserves scene structure during the diffusion process and is much simpler for learning the geometry information, while 3D Gaussian deformation often introduces artifacts at occlusion boundaries due to data scarcity. This prediction fidelity directly impacts action generation -- accurate future state prediction enables better trajectory planning, particularly for tasks involving precise object interactions.

\begin{figure}[!t]
    \includegraphics[width=\textwidth]{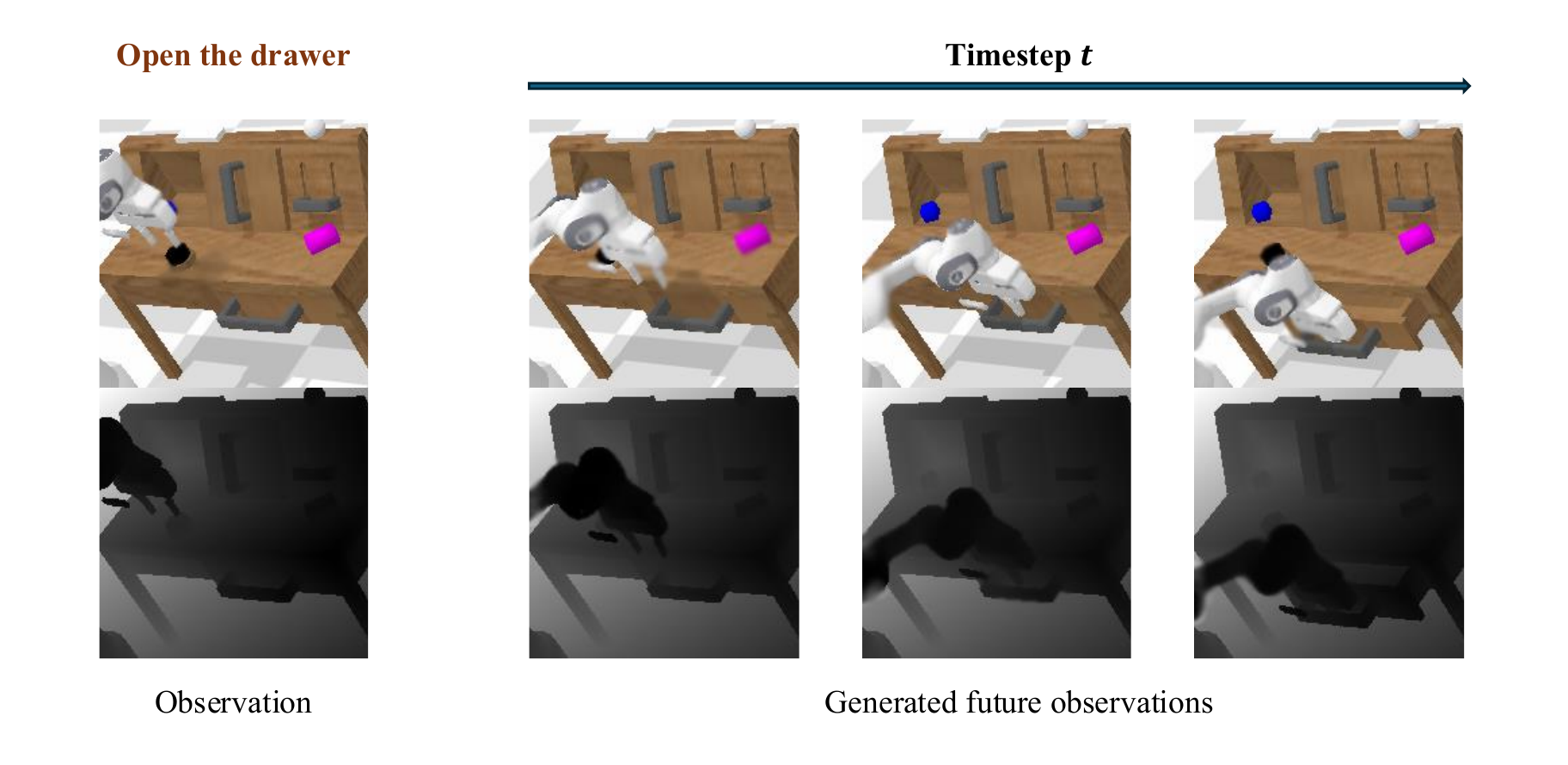}
    \vspace{-20pt}
    \caption{\textbf{Future observation prediction on Calvin~\citep{Mees2021CALVINAB}.}}
    \label{fig:calvin_vis}
    \vspace{-10pt}
\end{figure}

\textbf{Visualization.} Besides the visualization shown in Fig.~\ref{fig:rec} on RLBench, we also provide the future observations prediction results on Calvin. Fig.~\ref{fig:calvin_vis} illustrates the generated future observations conditioned on the current scene and the inferred latent action $\mathbf{z}^t$ for the ``Open the drawer'' task. 
Our model produces highly accurate and temporally consistent predictions in both RGB and depth modalities. 
The generated frames faithfully capture the geometric displacement of the drawer and the manipulator trajectory over multiple timesteps, showing smooth and physically plausible motion progression. 
Importantly, the predicted depth maps remain well aligned with the RGB predictions, indicating that the model preserves 3D scene structure rather than merely hallucinating pixel-level appearance. This coherence highlights that the latent action representation encodes the causal effect of actions in 3D space, enabling the diffusion model to generate consistent future trajectories.

\section{Real-world experiments}
\label{app:real}

\subsection{Dataset composition}
\begin{table}[htbp]
\centering
\begin{tabular}{lll}
\toprule
\textbf{Task} & \textbf{Variation Type} & \textbf{Language Description Example} \\
\midrule
Place dish & color, count & ``Place the --- dish on the --- tablecloth'' \\
Clean cups & color, count,  placement & ``Put the --- cup into the --- basket'' \\
Stack cups & color, count, placement & ``Stack the --- cup on the --- cup'' \\
Stack cubes & color, count & ``Stack the --- cubes'' \\
Put cups on shelf & placement & ``Put the --- cup on the shelf next to the --- cup'' \\
Place cube & color, count & ``Place the --- block on the --- plate'' \\
\bottomrule
\end{tabular}
\caption{Dataset composition of 6 manipulation tasks in real robot experiments.}
\label{tab:real_robot_tasks}
\end{table}

\begin{figure}[!t]
    \includegraphics[width=\textwidth]{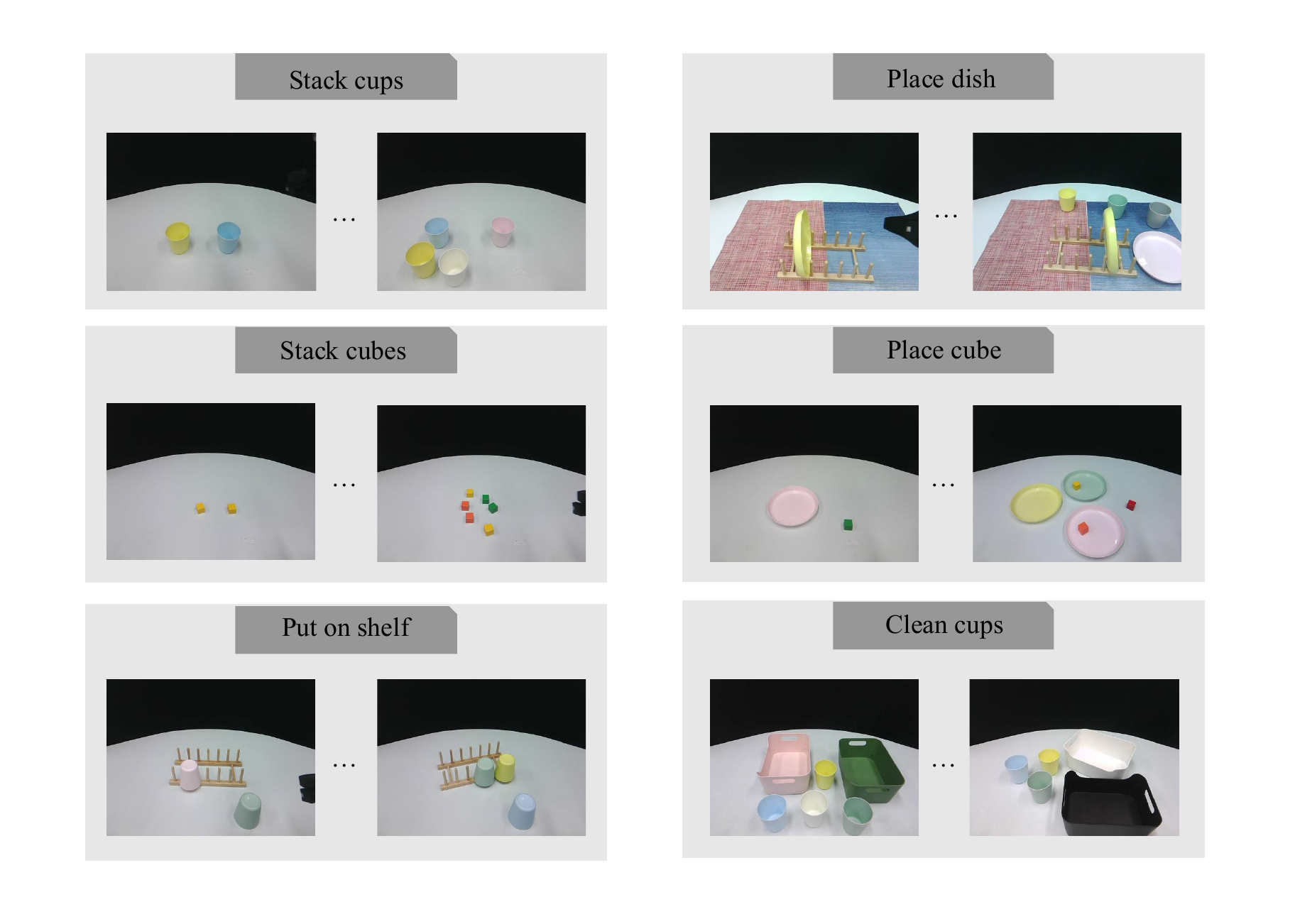}
    \vspace{-20pt}
    \caption{\textbf{Visualization of the real-world experiment setup.}}
    \label{fig:real_setting}
    \vspace{-10pt}
\end{figure}

We evaluate our approach on six real-world manipulation tasks, each incorporating controlled variations in object placement, color, and count. This design introduces perceptual and spatial diversity, challenging the agent to generalize across visually distinct yet semantically similar scenarios. The visualization of our settings is shown in Fig.~\ref{fig:real_setting}.

\subsection{More experimental results}
\begin{figure}[!t]
    \includegraphics[width=\textwidth]{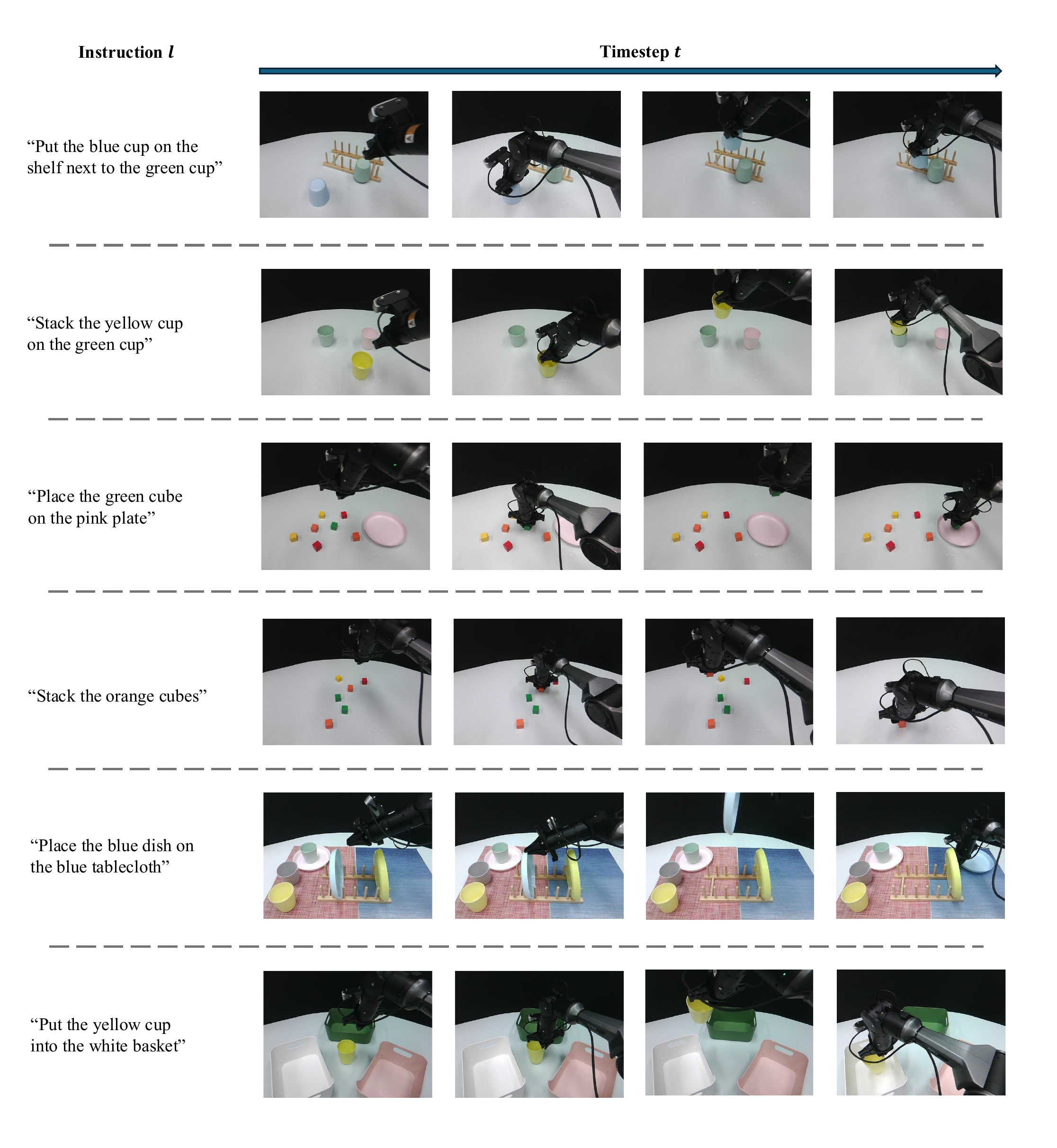}
    \vspace{-20pt}
    \caption{\textbf{More qualitative results on real-world experiments.}}
    \label{fig:real_demo}
    \vspace{-10pt}
\end{figure}

\paragraph{Real-World Experiments.}
We deploy our trained policy on a real Franka Panda manipulator and evaluate it on six long-horizon tabletop manipulation tasks: 
(1) ``Put the blue cup on the shelf next to the green cup'', 
(2) ``Stack the yellow cup on the green cup'', 
(3) ``Place the green cube on the pink plate'', 
(4) ``Stack the orange cubes'', 
(5) ``Place the blue dish on the blue tablecloth'', and 
(6) ``Put the yellow cup into the white basket''. 
As shown in Fig.~\ref{fig:real_demo}, our policy successfully completes all six tasks with smooth and collision-free trajectories, demonstrating strong sim-to-real transfer. 
The robot consistently executes precise grasps, object placements, and stacking behaviors, even in cluttered and visually diverse scenes. 
These results confirm that the learned latent actions generalize to real-world execution and maintain their semantic meaning outside of the simulation domain.

\begin{figure}[!t]
    \includegraphics[width=\textwidth]{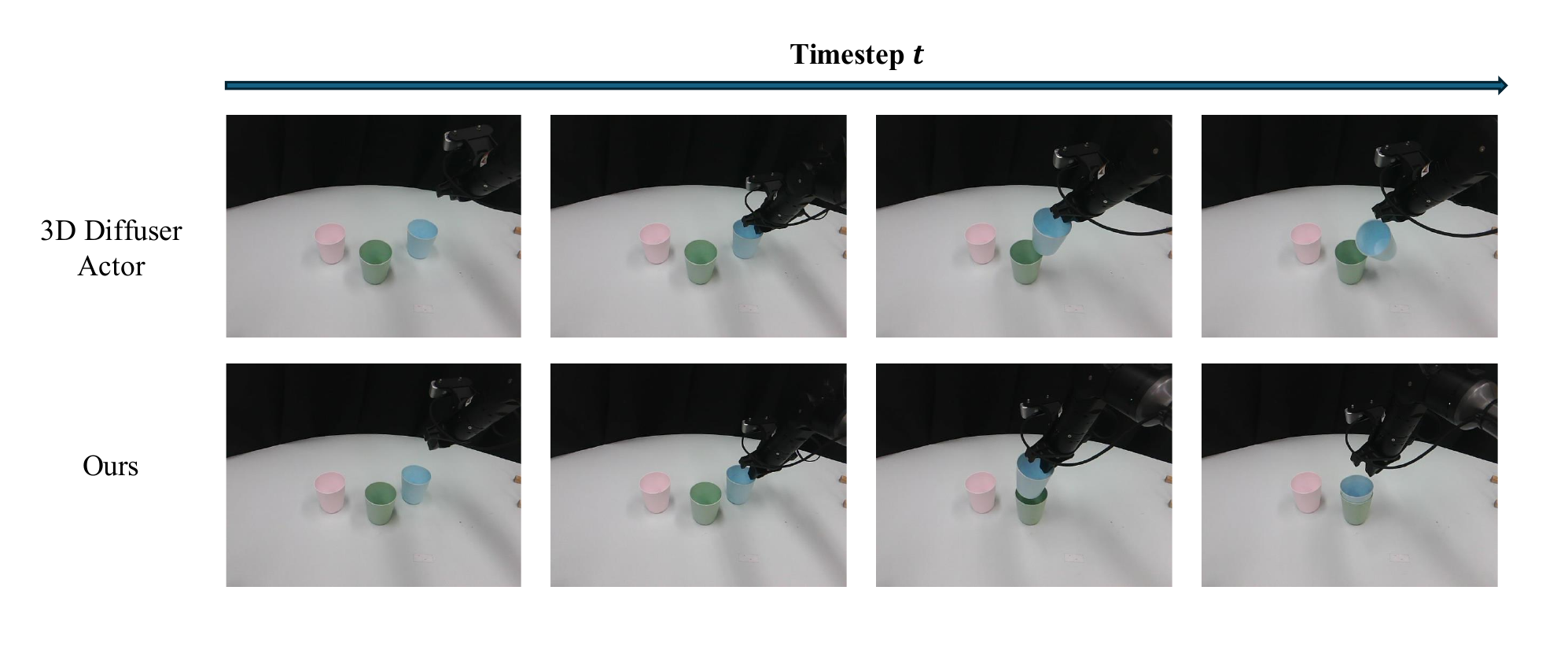}
    \vspace{-20pt}
    \caption{\textbf{Qualitative comparison in the real-world cluttered scene.}}
    \label{fig:real_vis}
    \vspace{-10pt}
\end{figure}

\textbf{Case study of precise action control.} Fig.~\ref{fig:real_vis} presents a qualitative comparison of action execution in a real-world cluttered scene, highlighting the advantage of our 4D modeling approach over a 3D static baseline. The top row (3D Diffuser Actor) demonstrates failure: although the robot successfully grasps and lifts the target cup, it misaligns during placement due to a lack of temporal dynamics and spatial reasoning — resulting in an unstable or incorrect stack. In contrast, our method (bottom row) leverages 4D spatio-temporal modeling to predict not only where but when and how to act, enabling precise control throughout the motion trajectory. As shown, our agent successfully stacks the cup with stable alignment, even under visual occlusion and object clutter. This illustrates that 4D-aware policy learning is critical for achieving reliable, fine-grained manipulation in dynamic physical environments — a capability absent in purely 3D state-based models.

\section{Implementation details}
\label{app:imple}

\begin{table}[htbp]
\centering
\begin{tabular}{lcc}
\toprule
 & \textbf{RLBench} & \textbf{CALVIN} \\
\midrule
\multicolumn{3}{l}{\textbf{Transformer}} \\
\quad image\_size & 256 & 200 \\
\quad embedding\_dim & 120 & 192 \\
\quad camera\_views & 1 & 2 \\
\quad FPS : \% of sampled tokens & 20\% & 33\% \\
\quad diffusion\_timestep & 100 & 25 \\

\midrule

\multicolumn{3}{l}{\textbf{Latent action}} \\
\quad Patch size & 16 & 16 \\
\quad Hidden size & 768 & 768 \\
\quad Codebook size & 64 & 64 \\
\quad Codebook dim & 32 & 32 \\
\midrule

\multicolumn{3}{l}{\textbf{Training}} \\
\quad batch\_size & 240 & 5400 \\
\quad learning\_rate & $1e^{-4}$ & $3e^{-4}$ \\
\quad weight\_decay & $5e^{-4}$ & $5e^{-3}$ \\
\quad total\_epochs & $1.6e^{4}$ & 90 \\
\quad optimizer & Adam & Adam \\

\bottomrule
\end{tabular}
\caption{Comparison of configurations between RLBench and CALVIN.}
\label{tab:hyper_param}
\end{table}
\subsection{Learning geometry-aware latent actions.}
To learn geometry-aware latent actions, we employ two different branches of diffusion models to predict the future observations.

Given RGB-D observation $\mathbf{o}^t$, we convert it to a pointmap:
\[
\mathbf{P}^t = \text{BackProject}(\mathbf{o}^t) \in \mathbb{R}^{H \times W \times 3}.
\]

\textbf{Latent Encoding.}
Pointmaps are encoded by a 3D-aware VAE:
\[
\mathbf{h}_{\text{pm}}^{t'} = \psi^{\text{enc}}(\mathbf{P}^{t'}), \quad
\hat{\mathbf{P}}^{t'} = \psi^{\text{dec}}(\hat{\mathbf{h}}_{\text{pm}}^{t'}).
\]

The pointmap VAE is initialized by the RGB VAE. Before the latent action learning, we first finetune it on the video sequences of demonstration for a few epochs.

\textbf{Latent Action-Conditioned Future Prediction.}
We generate future latents with a conditional diffusion model adapted from DDPM~\cite{Ho2020DenoisingDP} {with motion latent $z^t$}:
\[
\hat{\mathbf{h}}_{\text{pm}}^{t+1:t+w} = 
\psi^{\text{diff}}\!\left(\mathbf{h}_{\text{pm}}^{t-w+1:t}, \mathbf{z}^t; \phi \right).
\]

At each diffusion step $k$, we add Gaussian noise:
\[
\mathbf{h}_k^{\text{pm}} = \sqrt{\alpha_k}\cdot\mathbf{h}_{\text{pm}}^{t+1:t+w} + 
\sqrt{1-\alpha_k}\cdot\epsilon, \quad
\epsilon \sim \mathcal{N}(0,I),
\]
and minimize the denoising objective:
\[
\mathcal{L}_{\text{diff}}^{\text{pm}} = 
\mathbb{E}_{k,\epsilon}\!
\left[
\|\epsilon - \epsilon_{\phi}(\mathbf{h}_k^{\text{pm}}, k, 
\mathbf{h}_{\text{pm}}^{t-w+1:t}, \mathbf{z}^t)\|^2
\right].
\]

\textbf{Joint RGB Prediction.}
An RGB branch predicts future appearance using the same architecture and conditioning:
\[
\mathcal{L}_{\text{diff}}^{\text{rgb}} =
\mathbb{E}_{k,\epsilon}
\left[
\|\epsilon - \epsilon_{\phi}^{\text{rgb}}(\mathbf{h}_k^{\text{rgb}}, k, 
\mathbf{h}_{\text{rgb}}^{t-w+1:t}, \mathbf{z}^t)\|^2
\right].
\]

\textbf{Vector-Quantized Latent Regularization.}
Latent actions are discretized using VQ-VAE loss:
\[
\mathcal{L}_{\text{vq}} = 
\|\text{sg}[\mathbf{f}^t] - \mathbf{c}\|_2^2 
+ \beta \|\mathbf{f}^t - \text{sg}[\mathbf{c}]\|_2^2.
\]

\textbf{Overall Objective.}
The total training loss is:
\[
\mathcal{L}_{\text{total}} = 
\mathcal{L}_{\text{diff}}^{\text{pm}} +
\mathcal{L}_{\text{diff}}^{\text{rgb}} +
\mathcal{L}_{\text{vq}}.
\]

Note that both two VAEs are all frozen during the training process.

\subsection{Action prediction with geometry-aware latent actions.}

The 3D denoising transformer $\epsilon_\theta$ is a multi-layer conditional diffusion model that iteratively refines noisy action sequences into executable robot trajectories. The model takes four types of input tokens: trajectory tokens, scene tokens, proprioception tokens, and latent action tokens.

\textbf{Token Embeddings.} 
We embed each noisy action $\mathbf{a}^{t+k}$ using a MLP, producing a $d_{\text{model}}=256$-dimensional trajectory token. Scene features are extracted using a frozen CLIP-ResNet50 encoder and lifted to 3D with depth information and camera intrinsics.
The proprioceptive state $\mathbf{c}^t$ is encoded with an MLP and added with the positional embedding of the end-effector pose.

\textbf{Transformer Backbone.}
The denoising network contains:
(i) a multi-head self-attention layer over all trajectory, scene, and proprioception tokens with rotary 3D positional encodings;
(ii) a cross-attention layer that attends from all tokens to the latent action embeddings $z_t$ to inject high-level motion priors; In order to decrease the computational
requirements, we subsample a
number of visual tokens using Farthest Point Sampling (FPS). The sampled visual tokens, proprioception tokens, and noisy position/rotation tokens attend to each other.

\textbf{Output Heads.}
The final transformer layer outputs refined trajectory tokens, which are decoded using two independent MLP heads:
One predicts the noise of position and 6D rotation, and the other predicts the gripper open/close state.

\textbf{3D relative attention.} We formulate the detailed attention as follows:

\[
a_{q,k} \propto x_q^\top M(p_q - p_k) x_k
\]
\begin{itemize}
  \item \( a_{q,k} \): attention weight between query token \( q \) and key token \( k \)
  \item \( x_q \): feature vector of the query token
  \item \( x_k \): feature vector of the key token
  \item \( p_q \): 3D position of the query token
  \item \( p_k \): 3D position of the key token
  \item \( M(p_q - p_k) \): matrix-valued function that depends only on the relative 3D position between query and key
\end{itemize}

\subsection{Hyper-parameters} 

The summary of used hyper-parameters for training/evaluating our model is described in Tab.~\ref{tab:hyper_param}.

\end{document}